\documentclass[final]{cvpr}

\usepackage{times}
\usepackage{epsfig}
\usepackage{graphicx}
\usepackage{amsmath}
\usepackage{amssymb}

\usepackage[pagebackref=true,breaklinks=true,colorlinks,bookmarks=false]{hyperref}

\usepackage{url}
\usepackage{tabulary}

\usepackage{multirow}
\usepackage[normalem]{ulem}
\usepackage[english]{babel}

\usepackage[utf8]{inputenc} 
\usepackage[T1]{fontenc}    
\usepackage{hyperref}       
\usepackage{booktabs}       
\usepackage{amsfonts}       
\usepackage{nicefrac}       
\usepackage{microtype}      
\usepackage{mathtools}

\usepackage{multirow}
\usepackage{float}
\usepackage{setspace}
\usepackage{array}
\usepackage{tabularx}
\usepackage{calc}
\usepackage{makecell}
\usepackage{subfig}
\usepackage[dvipsnames]{xcolor}

\def\confYear{CVPR 2021}
\begin{document}

\title{Bottleneck Transformers for Visual Recognition}

\author{
Aravind Srinivas$^1$
\enspace
Tsung-Yi Lin$^2$
\enspace
Niki Parmar$^2$
\enspace
Jonathon Shlens$^2$
\enspace
Pieter Abbeel$^1$
\enspace
Ashish Vaswani$^2$
\\
$^1$UC Berkeley
\qquad
$^2$Google Research\\
{\tt\small \{aravind\}@cs.berkeley.edu}
}

\maketitle

\newcommand{\apbbox}[1]{AP$^\text{bb}_\text{#1}$}
\newcommand{\apmask}[1]{AP$^\text{mk}_\text{#1}$}
\newcommand{\cgap}[2]{\fontsize{6pt}{1em}\selectfont{(${#1}${#2})}}

\newcommand{\ty}[1]{{\color{cyan}ty: {#1}}}

\definecolor{myred}{cmyk}{0, 0.7808, 0.4429, 0.1412}

\begin{abstract}
\vspace{-.5em}
We present BoTNet, a conceptually simple yet powerful backbone architecture that incorporates self-attention for multiple computer vision tasks including image classification, object detection and  instance segmentation. By just replacing the spatial convolutions with global self-attention in the final three bottleneck blocks of a ResNet and no other changes, our approach improves upon the baselines significantly on instance segmentation and object detection while also reducing the parameters, with minimal overhead in latency. Through the design of BoTNet, we also point out how ResNet bottleneck blocks with self-attention can be viewed as Transformer blocks. Without any bells and whistles, BoTNet achieves {\textbf{44.4}}\% Mask AP and {\textbf{49.7}}\% Box AP on the COCO Instance Segmentation benchmark using the Mask R-CNN framework; surpassing the previous best published single model and single scale results of ResNeSt~\cite{zhang2020resnest} evaluated on the COCO validation set. Finally, we present a simple adaptation of the BoTNet design for image classification, resulting in models that achieve a strong performance of {\textbf{84.7}}\% top-1 accuracy on the ImageNet benchmark while being up to {\bf 1.64x} faster in ``compute''\footnote{Forward and backward propagation for batch size 32} time than the popular EfficientNet models on TPU-v3 hardware. We hope our simple and effective approach will serve as a strong baseline for future research in self-attention models for vision.
\end{abstract}

\section{Introduction}

Deep convolutional backbone architectures~\cite{krizhevsky2012, simonyan2014very, he2016deep, xie2017aggregated, tan2019efficientnet} have enabled significant progress in image classification~\cite{russakovsky2015imagenet}, object detection~\cite{everingham2010pascal, lin2014microsoft, girshick2014rich, girshick2015fast, ren2015faster}, instance segmentation~\cite{hariharan2014simultaneous, dai2016instance, he2017mask}. Most landmark backbone architectures~\cite{krizhevsky2012, simonyan2014very, he2016deep} use multiple layers of $3 \times 3 $ convolutions.

While the convolution operation can effectively capture local information, vision tasks such as object detection, instance segmentation, keypoint detection require modeling long range dependencies. For example, in instance segmentation, being able to collect and associate scene information from a large neighborhood can be useful in learning relationships across objects~\cite{hu2018relation}. In order to globally aggregate the locally captured filter responses, convolution based architectures require stacking multiple layers~\cite{simonyan2014very, he2016deep}. Although stacking more layers indeed improves the performance of these backbones~\cite{zhang2020resnest}, an explicit mechanism to model global (non-local) dependencies could be a more powerful and scalable solution without requiring as many layers.

\begin{figure}[t]
\centering
\includegraphics[scale=.25]{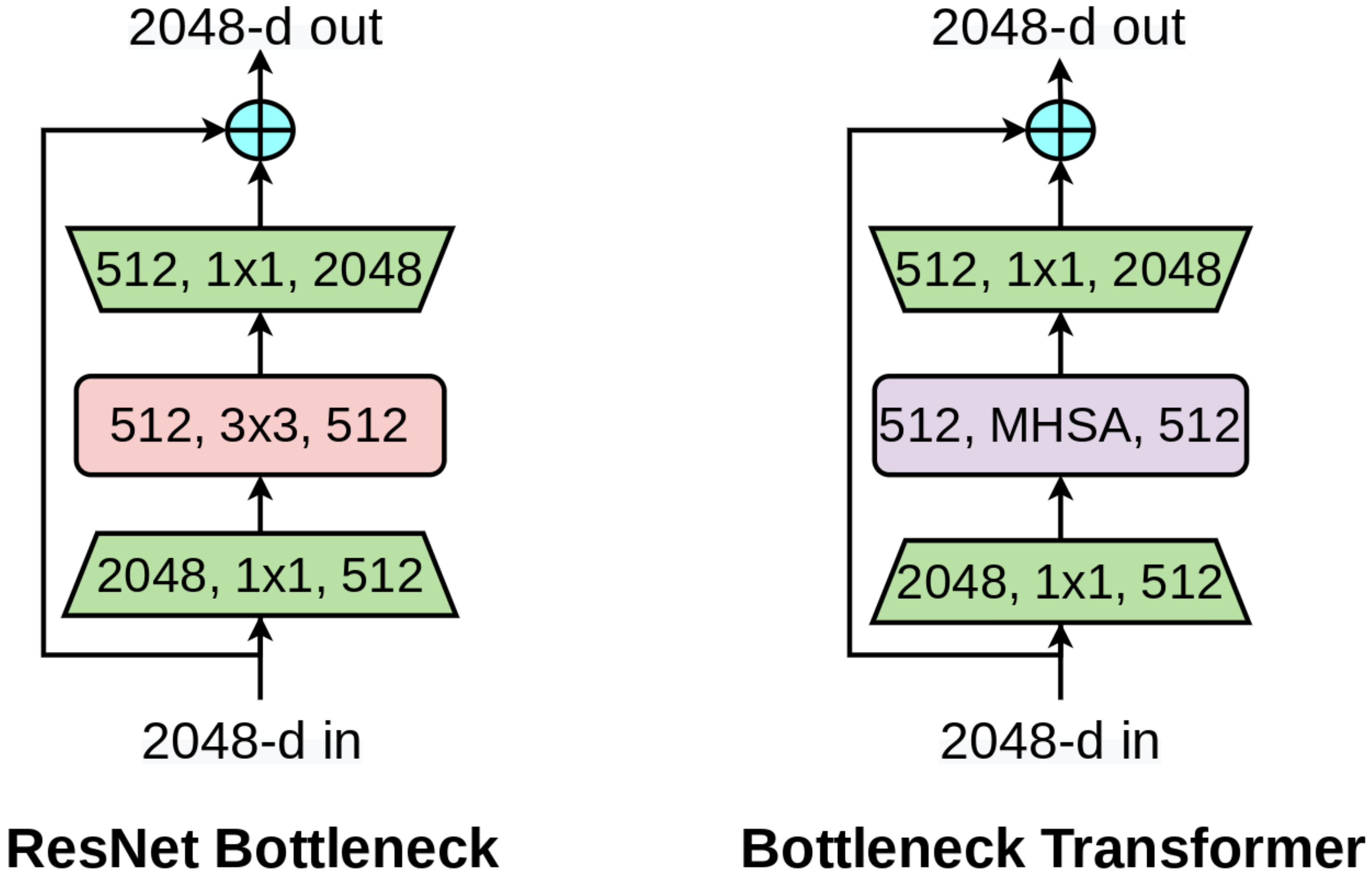}

\caption{{\bf Left:} A ResNet Bottleneck Block, {\bf Right:} A Bottleneck Transformer (BoT) block. The only difference is the replacement of the spatial $3\times 3$ convolution layer with Multi-Head Self-Attention (MHSA). The structure of the self-attention layer is described in Figure \ref{fig:alephdiagram}.}
\label{fig:teaser}
\end{figure}

Modeling long-range dependencies is critical to natural language processing (NLP) tasks as well. Self-attention is a computational primitive~\cite{vaswani2017attention} that implements pairwise entity interactions with a content-based addressing mechanism, thereby learning a rich hierarchy of associative features across long sequences. This has now become a standard tool in the form of Transformer~\cite{vaswani2017attention} blocks in NLP with prominent examples being GPT~\cite{radford2019language, brown2020language} and BERT~\cite{devlin2018bert, liu2019roberta} models.

\begin{figure*}[!ht]%
    \centering
    \includegraphics[scale=0.42]{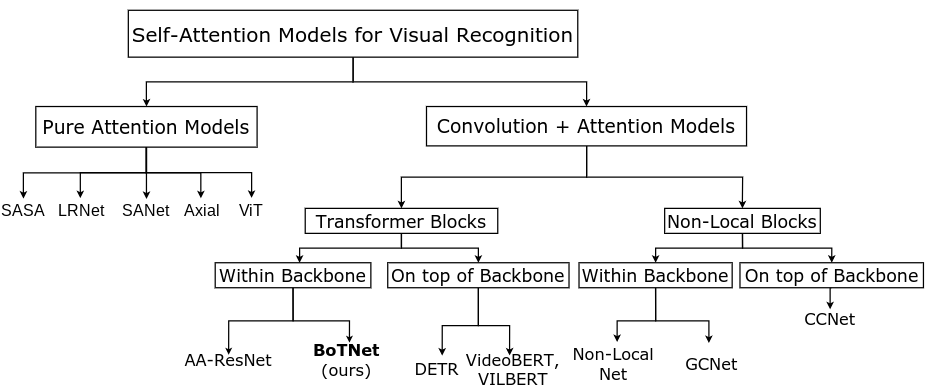}
    \caption{A taxonomy of deep learning architectures using self-attention for visual recognition. Our proposed architecture BoTNet is a hybrid model that uses both convolutions and self-attention. The specific implementation of self-attention could either resemble a Transformer block~\cite{vaswani2017attention} or a Non-Local block~\cite{wang2018non} (difference highlighted in Figure \ref{fig:alephdiagram}). BoTNet is different from architectures such as DETR~\cite{carion2020end}, VideoBERT~\cite{sun2019videobert}, VILBERT~\cite{lu2019vilbert}, CCNet~\cite{huang2019ccnet}, etc by employing self-attention within the backbone architecture, in contrast to using them outside the backbone architecture. Being a hybrid model, BoTNet differs from pure attention models such as SASA~\cite{ramachandran2019stand}, LRNet~\cite{hu2019local}, SANet~\cite{zhao2020exploring}, Axial-SASA~\cite{ho2019axial, wang2020axial} and ViT~\cite{dosovitskiy2020image}. AA-ResNet~\cite{bello2019attention} also attempted to replace a fraction of spatial convolution channels with self-attention.}%
    \label{fig:taxonomy}%
\end{figure*}

A simple approach to using self-attention in vision is to replace spatial convolutional layers with the multi-head self-attention (MHSA) layer proposed in the Transformer~\cite{vaswani2017attention} (Figure \ref{fig:teaser}). This approach has seen progress on two {seemingly different} approaches in the recent past. On the one hand, we have models such as SASA~\cite{ramachandran2019stand}, AACN~\cite{bello2019attention}, SANet~\cite{zhao2020exploring}, Axial-SASA~\cite{wang2020axial}, etc that propose to replace spatial convolutions in ResNet botleneck blocks~\cite{he2016deep} with different forms of self-attention (local, global, vector, axial, etc). On the other hand, we have the Vision Transformer (ViT)~\cite{dosovitskiy2020image}, that proposes to stack Transformer blocks~\cite{vaswani2017attention} operating on linear projections of non-overlapping patches. It may appear that these approaches present two different classes of architectures. We point out that it is {\it not the case}. Rather, ResNet botteneck blocks with the MHSA layer can be viewed as Transformer blocks with a bottleneck structure, modulo minor differences such as the residual connections, choice of normalization layers, etc. (Figure \ref{fig:equivalence}). Given this equivalence, we call ResNet bottleneck blocks with the MHSA layer as {\it Bottleneck Transformer} (BoT) blocks. 

Here are a few challenges when using self-attention in vision: (1) Image sizes are much larger ($1024 \times 1024$) in object detection and instance segmentation compared to image classification ($224\times224$). (2) The memory and computation for self-attention scale quadratically with spatial dimensions~\cite{tay2020efficient}, causing overheads for training and inference. 

To overcome these challenges, we consider the following design: (1) Use convolutions to {\it efficiently} learn {\it abstract} and {\it low resolution} featuremaps from large images; (2) Use global ({\it all2all}) self-attention to process and aggregate the information contained in the featuremaps captured by convolutions. Such a hybrid design~\cite{bello2019attention} (1) uses existing and well optimized primitives for both convolutions and all2all self-attention; (2) can deal with large images efficiently by having convolutions do the spatial downsampling and letting attention work on smaller resolutions. Here is a simple practical instantiation of this hybrid design: Replace {\it only} the final three bottleneck blocks of a ResNet with BoT blocks {\it without any other changes}. Or in other words, take a ResNet and only replace the final three $3\times3$ convolutions with MHSA layers (Fig \ref{fig:teaser}, Table \ref{tab:arch}). This simple change improves the mask AP by 1.2\% on the COCO instance segmentation benchmark~\cite{lin2014microsoft} over our canonical baseline that uses ResNet-50 in the Mask R-CNN framework~\cite{he2017mask} with {\it no hyperparameter differences} and minimal overheads for training and inference.
Moving forward, we call this simple instantiation as BoTNet given its connections to the Transformer through the BoT blocks. While we note that there is no novelty in its construction, we believe the simplicity and performance make it a useful reference backbone architecture that is worth studying.

Using BoTNet, we demonstrate significantly improved results on instance segmentation \emph{without any bells and whistles} such as Cascade R-CNN~\cite{cai2018cascade}, FPN changes~\cite{liu2018path, ghiasi2019fpn, liu2019cbnet, tan2019efficientdet}, hyperparameter changes~\cite{tan2019efficientnet}, etc. A few key results from BoTNet are: 
(1) Performance gains across various training configurations (Section \ref{sec:gains}), data augmentations (Section \ref{sec:jit}) and ResNet family backbones (Section \ref{sec:backbones}); (2) Significant boost from BoTNet on small objects (+2.4 Mask AP and +2.6 Box AP) (Appendix); (3) Performance gains over Non-Local layers (Section \ref{sec:nonlocal}); (4) Gains that scale well with larger images resulting in {\bf 44.4\%} mask AP, competitive with state-of-the-art performance among entries that only study backbone architectures with modest training schedules (up to 72 epochs) and no extra data or augmentations.\footnote{SoTA is based on \url{https://paperswithcode.com/sota/instance-segmentation-on-coco-minival}.}. 


\newpage

Lastly, we scale BoTNets, taking inspiration from the training and scaling strategies in~\cite{tan2019efficientnet, ramachandran2019stand, lee2020compounding, ridnik2021tresnet, radosavovic2020designing, zhang2020resnest, bello2021revisit}, after noting that BoTNets do not provide substantial gains in a smaller scale training regime. We design a family of BoTNet models that achieve up to {\bf 84.7\%} top-1 accuracy on the ImageNet validation set, while being upto {\bf 1.64x} faster than the popular EfficientNet models in terms of {\it compute} time on TPU-v3 hardware. By providing {\emph{strong results}} through BoTNet, we hope that self-attention becomes a widely used primitive in future vision architectures.

\section{Related Work}
\label{sec:related_work}

A taxonomy of deep learning architectures that employ self-attention for vision is presented in Figure \ref{fig:taxonomy}. In this section, we focus on: (1) Transformer vs BoTNet; (2) DETR vs BoTNet; (3) Non-Local vs BoTNet. 

\begin{figure}[ht]%
    \centering
    \includegraphics[scale=0.35]{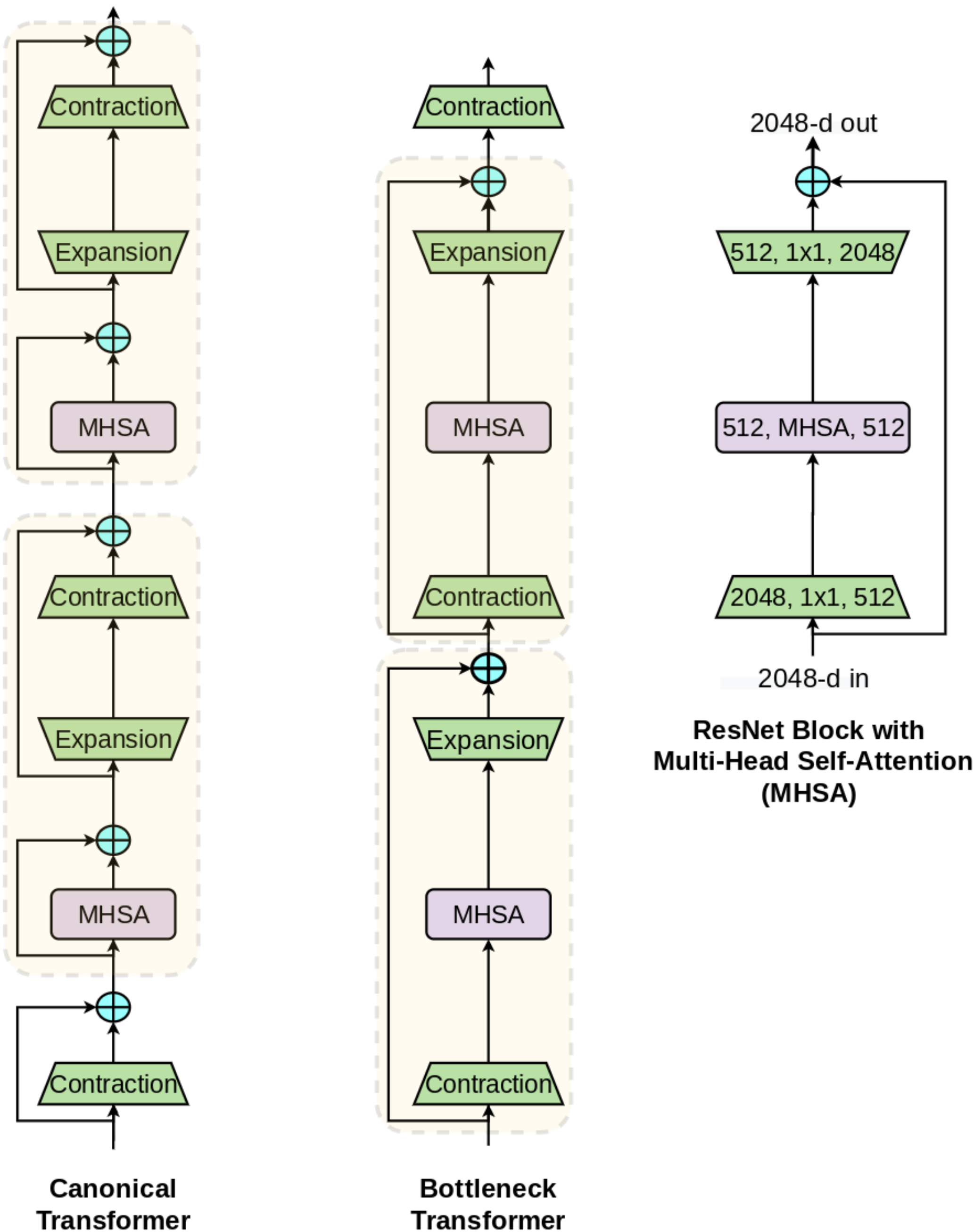}
    \caption{{\bf Left:} Canonical view of the Transformer with the boundaries depicting the definition of a Transformer block as described in Vaswani et. al~\cite{vaswani2017attention}. {\bf Middle:} Bottleneck view of the Transformer with boundaries depicting what we define as the Bottleneck Transformer (BoT) block in this work. The architectural structure that already exists in the Transformer can be interpreted a ResNet bottleneck block~\cite{he2016deep} with Multi-Head Self-Attention (MHSA)~\cite{vaswani2017attention} with a different notion of block boundary as illustrated. {\bf Right:} An instantiation of the Bottleneck Transformer as a ResNet bottleneck block~\cite{he2016deep} with the difference from a canonical ResNet block being the replacement of $3 \times 3$ convolution with MHSA.}%
    \label{fig:equivalence}%
\end{figure}

{\bf Connection to the Transformer:} As the title of the paper suggests, one key message in this paper is that ResNet bottleneck blocks with Multi-Head Self-Attention (MHSA) layers can be viewed as Transformer blocks with a bottleneck structure. This is visually explained in Figure \ref{fig:equivalence} and we name this block as Bottleneck Transformer (BoT). We note that the architectural design of the BoT block is not our contribution. Rather, we point out the relationship between MHSA ResNet bottleneck blocks and the Transformer with the hope that it improves our understanding of architecture design spaces~\cite{radosavovic2019network, radosavovic2020designing} for self-attention in computer vision. There are still a few differences aside from the ones already visible in the figure (residual connections and block boundaries): (1) Normalization: Transformers use Layer Normalization~\cite{ba2016layer} while BoT blocks use Batch Normalization~\cite{ioffe2015batch} as is typical in ResNet bottleneck blocks~\cite{he2016deep}; (2) Non-Linearities: Transformers use one non-linearity in the FFN block, while the ResNet structure allows BoT block to use three non-linearities; (3) Output projections: The MHSA block in a Transformer contains an output projection while the MHSA layer (Fig \ref{fig:alephdiagram}) in a BoT block (Fig \ref{fig:teaser}) does not; (4) We use the SGD with momentum optimizer typically used in computer vision~\cite{he2016deep, he2017mask, girshick2018detectron} while Transformers are generally trained with the Adam optimizer~\cite{kingma2014adam, vaswani2017attention, carion2020end, dosovitskiy2020image}. 

{\bf Connection to DETR:} Detection Transformer (DETR) is a detection framework that uses a Transformer to implicitly perform region proposals and localization of objects instead of using an R-CNN~\cite{girshick2014rich, girshick2015fast, ren2015faster, he2017mask}. Both DETR and BoTNet attempt to use self-attention to improve the performance on object detection and instance (or panoptic) segmentation. The difference lies in the fact that DETR uses Transformer blocks outside the backbone architecture with the motivation to get rid of region proposals and non-maximal suppression for simplicity. On the other hand, the goal in BoTNet is to provide a backbone architecture that uses Transformer-like blocks for detection and instance segmentation. We are agnostic to the detection framework (be it DETR or R-CNN). We perform our experiments with the Mask~\cite{he2017mask} and Faster R-CNN~\cite{ren2015faster} systems and leave it for future work to integrate BoTNet as the backbone in the DETR framework. With visibly good gains on small objects in BoTNet, we believe there maybe an opportunity to address the lack of gain on small objects found in DETR, in future (refer to Appendix).

{\bf Connection to Non-Local Neural Nets:}\footnote{The replacement vs insertion contrast has previously been pointed out in AA-ResNet (Bello et. al)~\cite{bello2019attention}. The difference in our work is the complete replacement as opposed to fractional replacement in Bello et al.} Non-Local (NL) Nets~\cite{wang2018non} make a connection between the Transformer and the Non-Local-Means algorithm~\cite{buades2005non}. They insert NL blocks into the final one (or) two blockgroups (\texttt{c4,c5}) in a ResNet and improve the performance on video recognition and instance segmentation. Like NL-Nets~\cite{wang2018non, cao2019gcnet}, BoTNet is a hybrid design using convolutions and global self-attention. (1) Three differences between a NL layer and a MHSA layer (illustrated in Figure \ref{fig:alephdiagram}): use of multiple heads, value projection and position encodings in MHSA; (2) NL blocks use a bottleneck with channel factor reduction of $2$ (instead of $4$ in BoT blocks which adopt the ResNet structure); (3) NL blocks are {\it inserted} as {\it additional} blocks into a ResNet backbone as opposed to {\it replacing} existing convolutional blocks as done by BoTNet.  Section \ref{sec:nonlocal} offers a comparison between BoTNet, NLNet as well as a NL-like version of BoTNet where we {\it insert} BoT blocks in the same manner as NL blocks instead of replacing.
\section{Method}
\label{sec:method}
\newcommand{\blockb}[3]{\multirow{3}{*}{
\(\left[
\begin{array}{l}
\text{1$\times$1, #2}\\
[-.1em] \text{3$\times$3, #2}\\
[-.1em] \text{1$\times$1, #1}
\end{array}\right]\)$\times$#3}
}

\newcommand{\blockBoT}[3]{\multirow{3}{*}{
\(\left[
\begin{array}{c}
\text{1$\times$1, #2}\\
[-.1em] \text{\texttt{\textcolor{myred}{MHSA}}, #2}\\
[-.1em] \text{1$\times$1, #1}
\end{array}\right]\)$\times$#3}
}

\newcommand{\blockx}[3]{\multirow{3}{*}{
\(\left[
\begin{array}{l}
\text{1$\times$1, #2}\\
[-.1em] \text{3$\times$3, #2, $C$=32}\\
[-.1em] \text{1$\times$1, #1}\\
\end{array}\right]\)$\times$#3}
}

\newcolumntype{x}[1]{>\centering p{#1pt}}
\newcommand{\ft}[1]{\fontsize{#1pt}{1em}\selectfont}
\renewcommand\arraystretch{1.25}
\setlength{\tabcolsep}{1.2pt}
\begin{table}[ht]
\begin{center}
\footnotesize
\begin{tabular}{c|c|x{100}|c}
\hline
 stage & output & ResNet-50 & \textbf{BoTNet-50} \\
\hline
\texttt{c1} & $512\times512$ & 7$\times$7, 64, stride 2 & 7$\times$7, 64, stride 2 \\
\hline
\multirow{4}{*}{\texttt{c2}} & \multirow{4}{*}{$256\times256$} & 3$\times$3 max pool, stride 2 & 3$\times$3 max pool, stride 2 \\\cline{3-4}
  &  &  \blockb{256}{64}{3} & \blockb{256}{64}{3}\\
  &  &  & \\
  &  &  & \\
\hline

\multirow{3}{*}{\texttt{c3}} &  \multirow{3}{*}{$128\times128$} 
  & \blockb{512}{128}{4} &  \blockb{512}{128}{4}\\
  &  &  & \\
  &  &  & \\
\hline
\multirow{3}{*}{\texttt{c4}} & \multirow{3}{*}{$64\times64$} 
  & \blockb{1024}{256}{6} & \blockb{1024}{256}{6}\\
  &  &  & \\
  &  &  & \\
\hline
\multirow{3}{*}{\texttt{c5}} & \multirow{3}{*}{$32\times32$} 
& \blockb{2048}{512}{3} & \blockBoT{2048}{512}{3}\\
  &  &  & \\
  &  &  & \\
\hline

\multicolumn{2}{c|}{\small \# params.} & \small \textbf{25.5}$\times$$10^6$  & \small \textbf{20.8}$\times$$10^6$ \\
\hline
\multicolumn{2}{c|}{\small M.Adds} & \small \textbf{85.4}$\times$$10^9$  & \small \textbf{102.98}$\times$$10^9$ \\
\hline
\multicolumn{2}{c|}{\small TPU steptime} & {\bf 786.5} ms & {\bf 1032.66} ms \\
\hline
\end{tabular}
\end{center}
\caption{Architecture of BoTNet-50 (BoT50): The only difference in BoT50 from ResNet-50 (R50) is the use of MHSA layer (Figure \ref{fig:alephdiagram}) in \texttt{c5}. For an input resolution of $1024 \times 1024$, the MHSA layer in the first block of \texttt{c5} operates on $64 \times 64$ while the remaining two operate on $32 \times 32$. We also report the parameters, multiply-adds (m. adds) and training time throughput (TPU-v3 steptime on a \texttt{v3-8} Cloud-TPU). BoT50 has only 1.2x more m.adds. than R50. The overhead in training throughout is 1.3x. BoT50 also has 1.2x {\it fewer} parameters than R50. While it may appear that it is simply the aspect of performing slightly more computations that might help BoT50 over the baseline, we show that it is not the case in Section \ref{sec:backbones}.}
\label{tab:arch}
\vspace{-.5em}
\end{table}

\begin{figure}[ht]%
    \centering
    \includegraphics[scale=0.3]{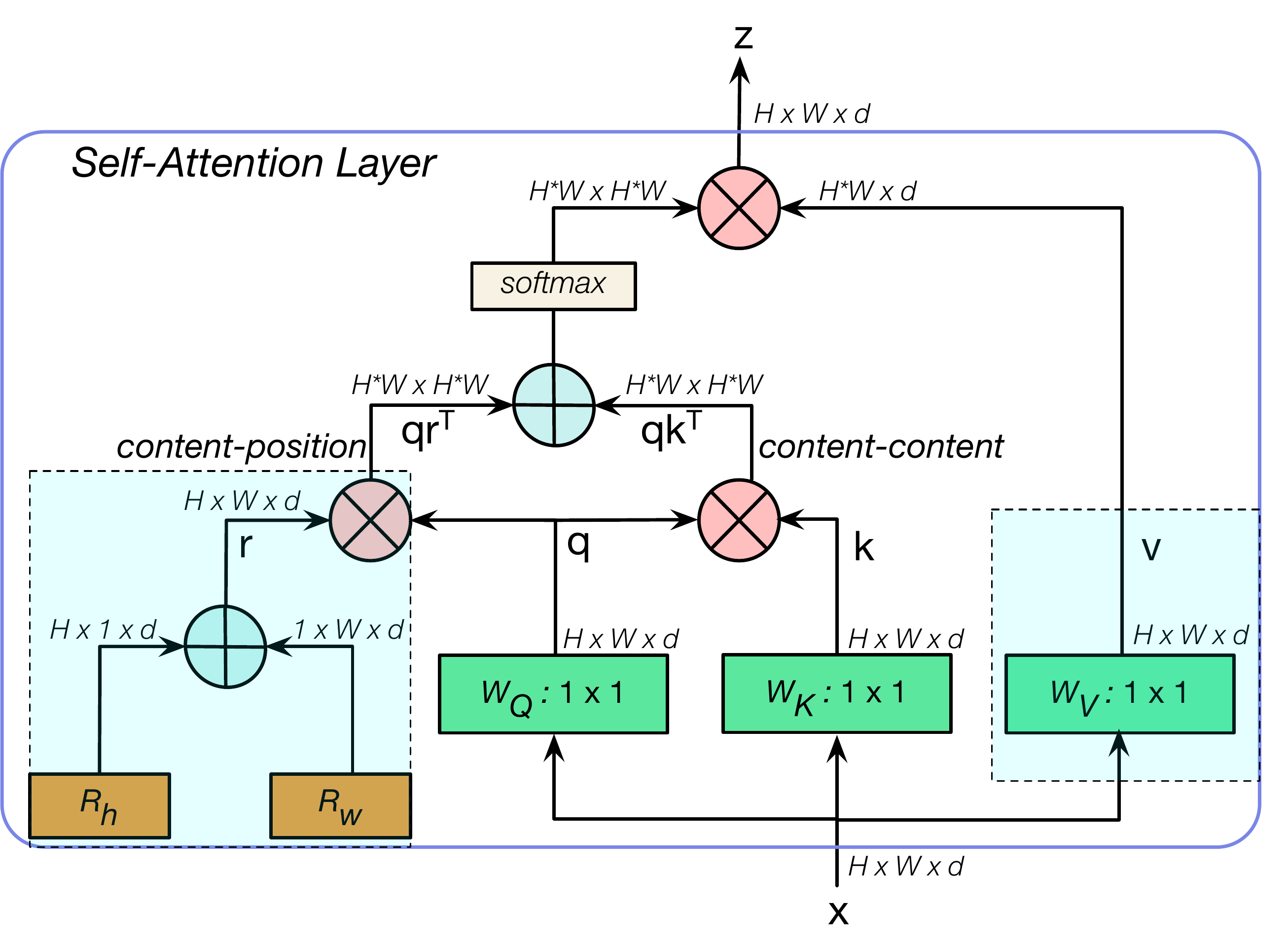}
    \caption{
     Multi-Head Self-Attention (MHSA) layer used in the BoT block. While we use 4 heads, we do not show them on the figure for simplicity. \texttt{all2all} attention is performed on a 2D featuremap with \emph{split} relative position encodings $R_h$ and $R_w$ for height and width respectively. The attention logits are $qk^T + qr^T$ where $q, k, r$ represent query, key and position encodings respectively (we use relative distance encodings~\cite{shaw2018self, bello2019attention, ramachandran2019stand}). $\bigoplus$ and $\bigotimes$ represent element wise sum and matrix multiplication respectively, while $1 \times 1$ represents a pointwise convolution. Along with the use of multiple heads, the highlighted blue boxes (\textcolor{cyan}{position encodings} and the \textcolor{cyan}{value projection} are the {\it only} three elements that are not present in the Non-Local Layer~\cite{wang2018non, xie2019feature}.}
    \label{fig:alephdiagram}%
\end{figure}

BoTNet by design is simple: replace the final three spatial ($3\times 3$) convolutions in a ResNet with Multi-Head Self-Attention (MHSA) layers that implement global (\emph{all2all}) self-attention over a 2D featuremap (Fig \ref{fig:alephdiagram}). A ResNet typically has 4 stages (or blockgroups) commonly referred to as \texttt{[c2,c3,c4,c5]} with strides \texttt{[4,8,16,32]} relative to the input image, respectively. Stacks \texttt{[c2,c3,c4,c5]} consist of multiple \emph{bottleneck} blocks with residual connections (e.g, R50 has \texttt{[3,4,6,3]} bottleneck blocks).

Approaches that use self-attention throughout the backbone~\cite{ramachandran2019stand, bello2019attention, zhao2020exploring, dosovitskiy2020image} are feasible for input resolutions ($224 \times 224$ (for classification) and $640 \times 640$ (for detection experiments in SASA~\cite{ramachandran2019stand})) considered in these papers.  Our goal is to use attention in more realistic settings of high performance instance segmentation models, where typically images of larger resolution ($1024 \times 1024$) are used. Considering that self-attention when performed globally across $n$ entities requires $ O(n^2 d)$ memory and computation~\cite{vaswani2017attention}, we believe that the simplest setting that adheres to the above factors would be to incorporate self-attention at the lowest resolution featuremaps in the backbone, ie, the residual blocks in the \texttt{c5} stack. The \texttt{c5} stack in a ResNet backbone typically uses 3 blocks with one spatial $3 \times 3$ convolution in each. Replacing them with MHSA layers forms the basis of the BoTNet architecture. The first block in \texttt{c5} uses a $3 \times 3$ convolution of stride $2$ while the other two use a stride of $1$. Since \texttt{all2all} attention is not a strided operation, we use a $2 \times 2$ average-pooling with a stride $2$ for the first BoT block. The BoTNet architecture is described in Table \ref{tab:arch} and the MHSA layer is presented in Figure \ref{fig:alephdiagram}. The strided version of the BoT block is presented in the Appendix.

\paragraph{Relative Position Encodings:} In order to make the attention operation {\it position aware}, Transformer based architectures typically make use of a position encoding~\cite{vaswani2017attention}. It has been observed lately that \emph{relative-distance-aware} position encodings~\cite{shaw2018self} are better suited for vision tasks~\cite{bello2019attention, ramachandran2019stand, zhao2020exploring}. This can be attributed to attention not only taking into account the content information but also relative distances between features at different locations, thereby, being able to effectively associate information across objects with positional awareness. In BoTNet, we adopt the 2D relative position self-attention implementation from ~\cite{ramachandran2019stand, bello2019attention}.

te\section{Experiments}
\label{sec:experiments}

We study the benefits of BoTNet for instance segmentation and object detection. We perform a thorough ablation study of various design choices through experiments on the COCO dataset~\cite{lin2014microsoft}. We report the standard COCO metrics including the \apbbox{} (averaged over IoU thresholds), \apbbox{50}, \apbbox{75}, \apmask{}; \apmask{50}, \apmask{75} for box and mask respectively. As is common practice these days, we train using the COCO \texttt{train} set and report results on the COCO \texttt{val} (or \texttt{minival}) set as followed in Detectron~\cite{girshick2018detectron}\footnote{\texttt{train} - 118K images, \texttt{val} - 5K images}. Our experiments are based on the Google Cloud TPU detection codebase\footnote{ \url{https://github.com/tensorflow/tpu/tree/master/models/official/detection}}. We run all the baselines and ablations with the same codebase. Unless explicitly specified, our training infrastructure uses \texttt{v3-8} Cloud-TPU which contains 8 cores with 16 GB memory per core. We train with the \texttt{bfloat16} precision and cross-replica batch normalization~\cite{ioffe2015batch, wu2018group, he2017mask, girshick2018detectron, peng2018megdet} using a batch size of 64.

\subsection{BoTNet improves over ResNet on COCO Instance Segmentation with Mask R-CNN}
\label{sec:gains}

\begin{table}[ht]\centering
 \small
 \setlength\tabcolsep{2.5pt}
 \begin{tabular}{l|c|c|c}
 \Xhline{1.0pt}
  \text{Backbone} & \text{epochs} & {\apbbox{}} &  {\apmask{}}  \\[2pt]
 \Xhline{1.0pt}

 R50   & 12 & 39.0   & 35.0   \\
 BoT50  & 12 &  39.4 (\textcolor{ForestGreen}{+ {\bf  0.4}}) &	35.3 (\textcolor{ForestGreen}{+ {\bf  0.3}})  \\
  \hline
 R50   & 24 & 41.2 & 36.9  \\
 BoT50 & 24 & 42.8 (\textcolor{ForestGreen}{+ {\bf 1.6}}) &  38.0 (\textcolor{ForestGreen}{+ {\bf 1.1}}) \\
  \hline
 R50 & 36 &	42.1  & 37.7 \\
 BoT50 & 36	& 43.6 (\textcolor{ForestGreen}{+ {\bf 1.5}}) &	38.9 (\textcolor{ForestGreen}{+ {\bf 1.2}}) \\
\hline
R50 & 72 & 42.8  &	37.9 \\
BoT50 & 72 & 43.7 (\textcolor{ForestGreen}{+ {\bf 0.9}}) & 38.7 (\textcolor{ForestGreen}{+ {\bf 0.8}}) \\
 \Xhline{1.0pt}
 \end{tabular}
 \vspace*{-0mm}
 \caption{{Comparing R50 and BoT50} under the 1x (12 epochs), 3x (36 epochs) and 6x (72 epochs) settings, trained with image resolution $1024 \times 1024$ and multi-scale jitter of $[0.8, 1.25]$.}
\label{tab:gains}
 \vspace{-2mm}
 \end{table}

We consider the simplest and most widely used setting: ResNet-50\footnote{We use the ResNet backbones pre-trained on ImageNet classification as is common practice. For BoTNet, the replacement layers are {\bf not} pre-trained but randomly initialized for simplicity; the remaining layers are initialized from a pre-trained ResNet.} backbone with FPN\footnote{FPN refers to Feature Pyramid Network~\cite{lin2017feature}. We use it in every experiment we report results on, and our FPN levels from 2 to 6 (\texttt{p2} to \texttt{p6}) similar to Detectron~\cite{girshick2018detectron}.}. We use images of resolution $1024 \times 1024$ with a multi-scale jitter of $[0.8, 1.25]$ (scaling the image dimension between $820$ and $1280$, in order to be consistent with the Detectron setting of using $800 \times 1300$). In this setting, we benchmark both the ResNet-50 (R50) and BoT ResNet-50 (BoT50) as the backbone architectures for multiple training schedules: {\bf 1x:} 12 epochs, {\bf 2x:} 24 epochs, {\bf 3x}: 36 epochs, {\bf 6x}: 72 epochs\footnote{1x, 2x, 3x and 6x convention is adopted from MoCo~\cite{he2019momentum}.}, all using the same hyperparameters for both the backbones across all the training schedules (Table \ref{tab:gains}). We clearly see that BoT50 is a significant improvement on top of R50 barring the 1x schedule (12 epochs). This suggests that BoT50 warrants longer training in order to show significant improvement over R50. We also see that the improvement from BoT50 in the 6x schedule (72 epochs) is worse than its improvement in the 3x schedule (32 epochs). This suggests that training much longer with the default scale jitter hurts. We address this by using a more aggressive scale jitter (Section \ref{sec:jit}).


\subsection{Scale Jitter helps BoTNet more than ResNet}
\label{sec:jit}

\begin{table}[ht]\centering
 \small
 \setlength\tabcolsep{2.0pt}
 \begin{tabular}{l|c|c|c}
\Xhline{1.0pt}
  \text{Backbone} & \texttt{jitter} & {\apbbox{}} &  {\apmask{}} \\[2pt]
 \Xhline{1.0pt}

R50   & $\left[0.8, 1.25\right]$ & 42.8 &	37.9  \\
BoT50  & $\left[0.8, 1.25\right]$ &  43.7 (\textcolor{ForestGreen}{+ {\bf 0.9}}) & 38.7 (\textcolor{ForestGreen}{+ {\bf 0.8}}) \\
\hline
R50   & $\left[0.5, 2.0\right]$ & 43.7 & 39.1 \\
BoT50  & $\left[0.5, 2.0\right]$ &  45.3 (\textcolor{ForestGreen}{+ {\bf 1.8}})  & 40.5 (\textcolor{ForestGreen}{+ {\bf 1.4}}) \\
\hline
R50   & $\left[0.1, 2.0\right]$ & 43.8 & 39.2  \\
BoT50  & $\left[0.1, 2.0\right]$ &  45.9 (\textcolor{ForestGreen}{+ {\bf 2.1}}) & 40.7 (\textcolor{ForestGreen}{+ {\bf 1.5}})	\\
\hline

\Xhline{1.0pt}
\end{tabular}
\vspace*{-0mm}
\caption{Comparing R50 and BoT50 under three settings of multi-scale jitter, all trained with image resolution $1024 \times 1024$ for 72 epochs (6x training schedule).}
\label{tab:jit}
\vspace{-2mm}
\end{table}

In Section \ref{sec:gains}, we saw that training much longer (72 epochs) reduced the gains for BoT50. One way to address this is to increase the amount of multi-scale jitter which has been known to improve the performance of detection and segmentation systems~\cite{du2019spinenet, ghiasi2020simple}. 
Table~\ref{tab:jit} shows that BoT50 is significantly better than R50 ({ + 2.1\%} on \apbbox{} and {+ 1.7\%} on \apmask{}) for multi-scale jitter of $[0.5, 2.0]$, while also showing significant gains ({ + 2.2\%} on \apbbox{} and {+ 1.6\%} on \apmask{}) for scale jitter of $[0.1, 2.0]$, suggesting that BoTNet (self-attention) benefits more from extra augmentations such as multi-scale jitter compared to ResNet (pure convolutions).

\subsection{Relative Position Encodings Boost Performance}
\label{sec:posenc}

BoTNet uses relative position encodings~\cite{shaw2018self}. We present an ablation for the use of relative position encodings by benchmarking the individual gains from content-content interaction ($qk^T$) and content-position interaction ($qr^T$) where $q, k, r$ represent the query, key and relative position encodings respectively. The ablations (Table \ref{tab:relpos}) are performed with the canonical setting\footnote{\texttt{res:1024x1024, 36 epochs (3x schedule), multi-scale jitter:[0.8, 1.25]}}. We see that the gains from $qr^T$ and $qk^T$ are complementary with $qr^T$ more important, ie, $qk^T$ standalone contributes to {0.6\%} \apbbox{} and {0.6\%} \apmask{} improvement over the R50 baseline, while $qr^T$ standalone contributes to {1.0\%} \apbbox{} and { 0.7 \%} \apmask{} improvement. When combined together ($qk^T + qr^T$), the gains on both \apbbox{} and \apmask{} are additive ({ 1.5\%} and {1.2\%} respectively). We also see that using absolute position encodings ($qr_{\textrm{abs}}^T$) does not provide as much gain as relative. This suggests that introducing relative position encodings into architectures like DETR~\cite{carion2020end} is an interesting direction for future work.

\begin{table}[H]\centering
 \small
 \setlength\tabcolsep{2.0pt}
 \begin{tabular}{l|c|c|c}
 \Xhline{1.0pt}
  \text{Backbone}  & \text{Att. Type} & {\apbbox{}} & {\apmask{}} \\[2pt]
 \Xhline{1.0pt}

 R50  & - & 42.1 &  37.7 \\
 BoT50 & $qk^T$ &  42.7 (\textcolor{ForestGreen}{+ {\bf 0.6}}) & 38.3 (\textcolor{ForestGreen}{+ {\bf 0.6}})	 \\
 BoT50 & $qr_{\textrm{relative}}^T$ & 43.1 (\textcolor{ForestGreen}{+ {\bf 1.0}}) & 38.4 (\textcolor{ForestGreen}{+ {\bf 0.7}})	\\
 BoT50 &  $qk^T + qr_{\textrm{relative}}^T$ & 43.6 (\textcolor{ForestGreen}{+ {\bf 1.5}}) & 38.9 (\textcolor{ForestGreen}{+ {\bf 1.2}}) \\
 BoT50 & $qk^T + qr_{\textrm{abs}}^T$ & 42.5 (\textcolor{ForestGreen}{+ {\bf 0.4}}) & 38.1 (\textcolor{ForestGreen}{+ {\bf 0.4}}) \\
 \hline

\Xhline{1.0pt}
\end{tabular}
\vspace*{-0mm}
\caption{Ablation for Relative Position Encoding: Gains from the two types of interactions in the MHSA layers, content-content ($qk^T$) and content-position ($qr^T$).}
\label{tab:relpos}
\vspace{-2mm}
\end{table}


\subsection{BoTNet improves backbones in ResNet Family}
\label{sec:backbones}
How well does the replacement setup of BoTNet work for other backbones in the ResNet family?
Table \ref{tab:backbones} presents the results for BoTNet with R50, R101, and R152. All these experiments use the canonical training setting (refer to footnote in \ref{sec:posenc}). These results demonstrate that BoTNet is applicable as a drop-in replacement for any ResNet backbone. Note that BoT50 is better than R101 ({+ 0.3\%} \apbbox{}, {\bf + 0.5\%} \apmask{}) while it is {competitive with R152} on \apmask{}. Replacing 3 spatial convolutions with \texttt{all2all} attention gives more improvement in the metrics compared to stacking 50 more layers of convolutions (R101), and is competitive with stacking 100 more layers (R152), supporting our initial hypothesis that {long-range dependencies are better captured through attention than stacking convolution layers.\footnote{Note that while one may argue that the improvements of BoT50 over R50 could be attributed to having 1.2x more M. Adds, BoT50 ($121 \times 10^9$  M.Adds) is also better than R101 ($162.99\times 10^{9}$ B  M. Adds and is competitive with R152 ($240.56 \times 10^{9}$ M. Adds) despite performing significantly less computation.}}

\begin{table}[h]\centering
\small
\setlength\tabcolsep{2.0pt}
\begin{tabular}{l|c|c}
\Xhline{1.0pt}
\text{Backbone}  & {\apbbox{}}  & {\apmask{}} \\[2pt]
\Xhline{1.0pt}

R50	& 42.1 & 	37.7  \\
BoT50 &	43.6 (\textcolor{ForestGreen}{+ {\bf 1.5}}) &	38.9 (\textcolor{ForestGreen}{+ {\bf 1.2}})  \\
\hline
R101	 &	43.3 &	38.4 \\
BoT101 & 45.5 (\textcolor{ForestGreen}{+ {\bf 2.2}}) &	40.4 (\textcolor{ForestGreen}{+ {\bf 2.0}})  \\
\hline
R152 &	44.2 &	39.1 \\
BoT152 &	46.0 (\textcolor{ForestGreen}{+ {\bf 1.8}}) &	40.6 (\textcolor{ForestGreen}{+ {\bf 1.5}})  \\

 \hline

 \Xhline{1.0pt}
 \end{tabular}
 \vspace*{-0mm}
 \caption{Comparing R50, R101, R152, BoT50, BoT101 and BoT152; all 6 setups using the canonical training schedule of 36 epochs, $1024\times 1024$ images, multi-scale jitter $\left[0.8, 1.25\right]$.}
 \label{tab:backbones}
 \vspace{-2mm}
 \end{table}

\subsection{BoTNet scales well with larger images}
\label{sec:botlarge}
We benchmark BoTNet as well as baseline ResNet when trained on $1280 \times 1280$ images in comparison to $1024 \times 1024$ using the best config: multi-scale jitter of $[0.1, 2.0]$ and training for 72 epochs. Results are presented in Tables \ref{tab:large} and \ref{tab:large_large}. Results in Table \ref{tab:large} suggest that BoTNet benefits from training on larger images for all of R50, R101 and R152. BoTNet trained on $1024 \times 1024$ (leave alone $1280 \times 1280$) is significantly better than baseline ResNet trained on $1280 \times 1280$. Further, BoT200 trained with $1280 \times 1280$ achieves a \apbbox{} of {\bf 49.7\%} and \apmask{} of {\bf 44.4\%}. We believe this result highlights the power of self-attention, in particular, because it has been achieved {without any bells and whistles} such as modified FPN~\cite{liu2018path, ghiasi2019fpn, du2019spinenet, tan2019efficientdet}, cascade RCNN~\cite{cai2018cascade}, etc. This result surpasses the previous best published single model single scale instance segmentation result from ResNeSt~\cite{zhang2020resnest} evaluated on the COCO \texttt{minival} (44.2\% \apmask{}).
\begin{table}[ht]\centering
 \small
 \setlength\tabcolsep{2.0pt}
 \begin{tabular}{l|c|c|c}
 \Xhline{1.0pt}
  \text{Backbone}  & \texttt{res} & {\apbbox{}} & {\apmask{}}  \\[2pt]
 \Xhline{1.0pt}

R50	& $1280$	& 44.0 & 39.5  \\
BoT50  & $1024$  &	45.9 (\textcolor{ForestGreen}{+ {\bf 1.9}}) &	40.7 (\textcolor{ForestGreen}{+ {\bf 1.2}})  \\
BoT50 & $1280$ &	{ 46.1} (\textcolor{ForestGreen}{+ {\bf 2.1}}) &	{ 41.2} (\textcolor{ForestGreen}{+ {\bf 1.8}}) \\

\hline

R101  & $1280$	& 46.4 &  41.2  \\
BoT101 & $1024$ & 47.4 (\textcolor{ForestGreen}{+ {\bf 1.0}}) & 42.0 (\textcolor{ForestGreen}{+ {\bf 0.8}}) \\
BoT101 & $1280$	& { 47.9} (\textcolor{ForestGreen}{+ {\bf 1.5}}) & { 42.4} (\textcolor{ForestGreen}{+ {\bf 1.2}}) \\
\hline

\Xhline{1.0pt}
\end{tabular}
\vspace*{-0mm}
\caption{
All the models are trained for 72 epochs with a multi-scale jitter of $[0.1, 2.0]$.}
\label{tab:large}
\vspace{-2mm}
\end{table}

\begin{table}[ht]\centering
 \small
 \setlength\tabcolsep{4.2pt}
 \begin{tabular}{l|c|c|c}
 \Xhline{1.0pt}
  \text{Backbone}  & \text{Change in backbone} & {\apbbox{}}  & {\apmask{}}  \\[2pt]
 \Xhline{1.0pt}
 
R50	& - &  42.1 &	37.7  \\
\hline
R50 + NL~\cite{wang2018non} &  + 1 NL block in \texttt{c4} & 43.1  &	38.4   \\
R50 + BoT (c4) & + 1 BoT block in \texttt{c4} &	43.7  &	38.9 \\
R50 + BoT (c4, c5) & + 2 BoT blocks in \texttt{c4,c5} &	44.9  & 39.7  \\
\hline
BoT50 & Replacement in \texttt{c5} & 43.6	& 38.9  \\

\hline

\Xhline{1.0pt}
\end{tabular}
\vspace*{-0mm}
\caption{Comparison between BoTNet and Non-Local (NL) Nets: All models trained for 36 epochs with image size $1024\times1024$, jitter $\left[0.8,1.25\right]$.}
\label{tab:nl}
\vspace{-2mm}
\end{table}

\begin{table}[ht]\centering
 \small
 \setlength\tabcolsep{5.5pt}
 \begin{tabular}{l|ccc|ccc}
 \Xhline{1.0pt}
  \text{Backbone}  & {\apbbox{}} & {\apbbox{50}} & {\apbbox{75}} & {\apmask{}} & {\apmask{50}} & {\apmask{75}} \\[2pt]
 \Xhline{1.0pt}
BoT152	 &  49.5 & 71.0 & 54.2 & 43.7 & 68.2 & 47.4 \\
BoT200  & {\bf 49.7} & {\bf 71.3} & {\bf 54.6} & {\bf 44.4} & {\bf 68.9} & {\bf 48.2} \\

\hline
\Xhline{1.0pt}
\end{tabular}
\vspace*{-0mm}
\caption{BoT152 and BoT200 trained for 72 epochs with a multi-scale jitter of $[0.1, 2.0]$.}
\label{tab:large_large}
\vspace{-2mm}
\end{table}


\subsection{Comparison with Non-Local Neural Networks}
\label{sec:nonlocal}

How does BoTNet compare to Non-Local Neural Networks? NL ops are {\it inserted} into the \texttt{c4} stack of a ResNet backbone between the pre-final and final bottleneck blocks. This {\it adds} more parameters to the model, whereas BoTNet ends up reducing the model parameters (Table \ref{tab:backbones}). In the NL mould, we add ablations where we introduce BoT block in the exact same manner as the NL block. We also run an ablation with the insertion of two BoT blocks, one each in the \texttt{c4,c5} stacks. Results are presented in Table \ref{tab:nl}. Adding a NL improves \apbbox{} by 1.0 and \apbbox{} by 0.7, while adding a BoT block gives +1.6 \apbbox{} and +1.2 \apmask{} showing that BoT block design is better than NL. Further, BoT-R50 (which replaces instead of adding new blocks) provides +1.5 \apbbox{} and + 1.2 \apmask{}, as good as adding another BoT block and better than adding one additional NL block.

\subsection{Image Classification on ImageNet}
\label{sec:imagenet_stuff}

\subsubsection{BoTNet-S1 architecture}

While we motivated the design of BoTNet for detection and segmentation, it is a natural question to ask whether the BoTNet architecture design also helps improve the image classification performance on the ImageNet~\cite{russakovsky2015imagenet} benchmark. Prior work~\cite{xie2019feature} has shown that {\it adding} Non-Local blocks to ResNets and training them using canonical settings does {\it not} provide substantial gains. We observe a similar finding for BoTNet-50 when contrasted with ResNet-50, with both models trained with the canonical hyperparameters for ImageNet~\cite{radosavovic2020designing}: 100 epochs, batch size 1024, weight decay 1e-4, standard ResNet data augmentation, cosine learning rate schedule (Table \ref{tab:imresults_canonical}). BoT50 does {\it not} provide significant gains over R50 on ImageNet though it does provide the benefit of reducing the parameters while maintaining comparable computation (M.Adds).

A simple method to fix this lack of gain is to take advantage of the image sizes typically used for image classification. In image classification, we often deal with much smaller image sizes ($224 \times 224$) compared to those used in object detection and segmentation ($1024 \times 1024)$. The featuremaps on which the BoT blocks operate are hence much smaller (e.g $14\times 14$, $7\times 7$) compared to those in instance segmentation and detection (e.g $64\times 64$, $32 \times 32$). With the same number of parameters, and, without a significant increase in computation, the BoTNet design in the \texttt{c5} blockgroup can be changed to uniformly use a stride of $1$ in all the final MHSA layers. We call this design as BoTNet-S1 (S1 to depict stride 1 in the final blockgroup). We note that this architecture is similar in design to the hybrid models explored in Vision Transformer (ViT)~\cite{dosovitskiy2020image} that use a ResNet up to stage \texttt{c4} prior to stacking Transformer blocks. The main difference between BoTNet-S1 and the hybrid ViT models lies in the use of BoT blocks as opposed to regular Transformer blocks (other differences being normalization layer, optimizer, etc as mentioned in the contrast to Transformer in Related Work (Sec. \ref{sec:related_work}). The architectural distinction amongst ResNet, BoTNet and BoTNet-S1, in the final blockgroup, is visually explained in the Appendix). The strided BoT block is visually explained in the Appendix.

\subsubsection{Evaluation in the standard training setting}
We first evaluate this design for the 100 epoch setting along with R50 and BoT50. We see that BoT-S1-50 improves on top of R50 by 0.9\% in the regular setting (Table \ref{tab:imresults_canonical}). This improvement does however come at the cost of more computation (m.adds). Nevertheless, the improvement is a promising signal for us to design models that scale well with larger images and improved training conditions that have become more commonly used since EfficientNets~\cite{tan2019efficientnet}.

\begin{table}[ht]\centering
 \small
 \setlength\tabcolsep{1.0pt}
 \begin{tabular}{l|c|c|c}
 \Xhline{1.0pt}
\text{Backbone}  &  M.Adds & Params & {top-1 acc.} \\[2pt]
\Xhline{1.0pt}

R50 &  3.86G & 25.5M & 76.8 \\
BoT50 &	3.79G	& 20.8M & 77.0 ({+{0.2}}) \\
BoT-S1-50 & 4.27G & 20.8M & 77.7 (\textcolor{ForestGreen}{+ {\bf 0.9}}) \\

\Xhline{1.0pt}
\end{tabular}
\vspace*{-0mm}
\caption{ImageNet results in regular training setting: 100 epochs, batch size 1024, weight decay 1e-4, standard ResNet augmentation, for all three models.}
\label{tab:imresults_canonical}
\vspace{-2mm}
\end{table}

\subsubsection{Effect of data augmentation and longer training}

We saw from our instance segmentation experiments that BoTNet and self-attention benefit more from regularization such as data augmentation (in the case of segmentation, increased multi-scale jitter) and longer training. It is natural to expect that the gains from BoT and BoT-S1 could improve when training under an improved setting: 200 epochs, batch size 4096, weight decay 8e-5, RandAugment (2 layers, magnitude 10), and label smoothing of 0.1. In line with our intuition, the gains are much more significant in this setting for both BoT50 (+ 0.6\%) and BoT-S1-50 (+ 1.4\%) compared to the baseline R50 (Table \ref{tab:imresults_improved}).

\begin{table}[ht]\centering
 \small
 \setlength\tabcolsep{1.0pt}
 \begin{tabular}{l|c|c}
 \Xhline{1.0pt}
\text{Backbone}  & {top-1 acc.} & {top-5 acc.} \\[2pt]
\Xhline{1.0pt}

R50 & 	77.7 &	93.9 \\
BoT50 &	78.3 (\textcolor{ForestGreen}{+ {\bf 0.6}}) &	94.2 (\textcolor{ForestGreen}{+ {\bf 0.3}}) \\
BoT-S1-50 &	79.1 (\textcolor{ForestGreen}{+ {\bf 1.4}}) & 94.4 (\textcolor{ForestGreen}{+ {\bf 0.5}}) \\

\Xhline{1.0pt}
\end{tabular}
\vspace*{-0mm}
\caption{ImageNet results in an improved training setting: 200 epochs, batch size 4096, weight decay 8e-5, RandAugment (2 layers, magnitude 10), and label smoothing of 0.1}
\label{tab:imresults_improved}
\vspace{-2mm}
\end{table}


\subsubsection{Scaling BoTNets}




\begin{figure}[ht]%
    \centering
    \includegraphics[scale=0.19]{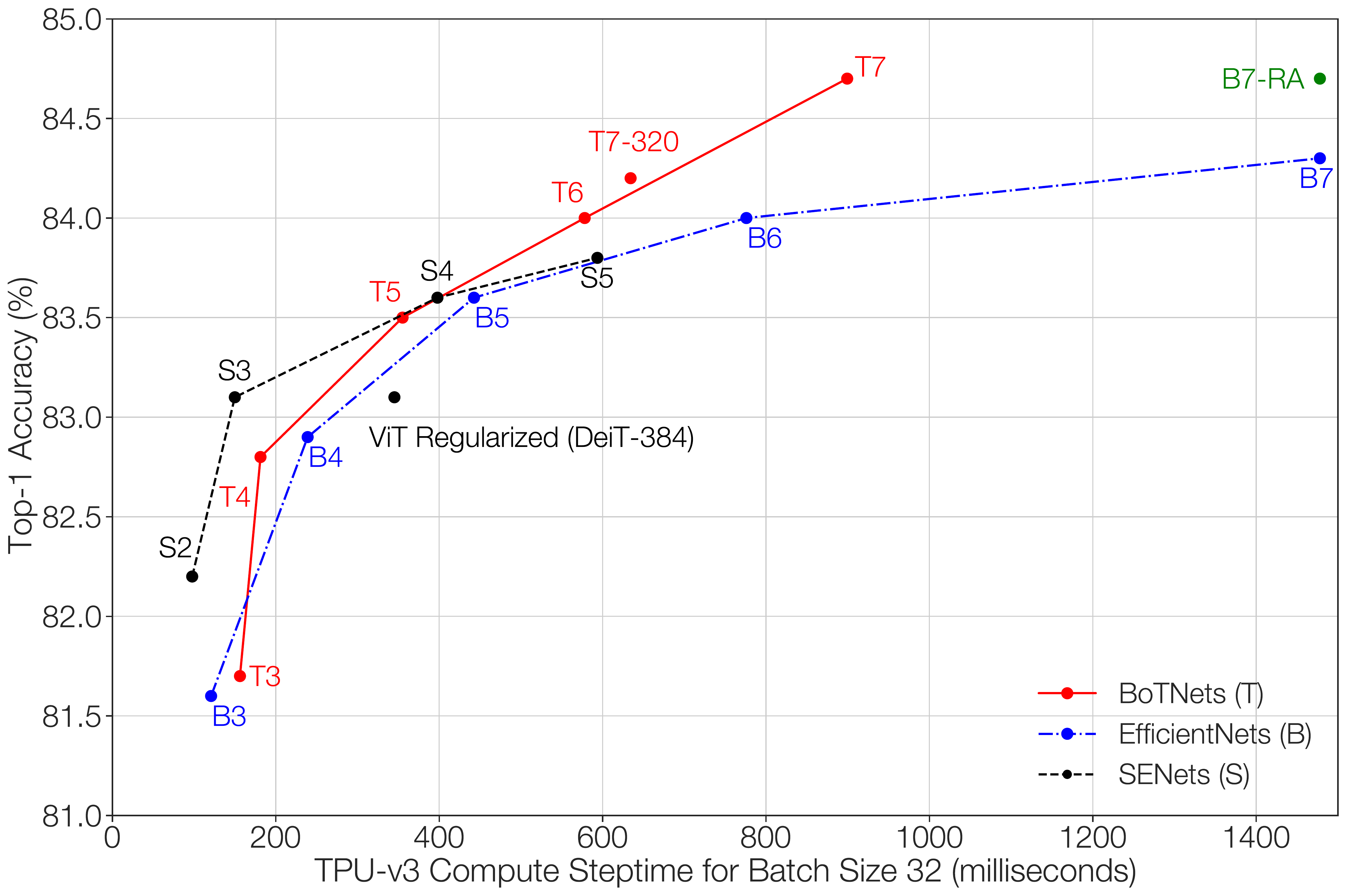}
    \caption{All backbones along with ViT and DeiT summarized in the form of scatter-plot and Pareto curves. SENets and BoTNets were trained while the accuracy of other models have been reported from corresponding papers.}%
    \label{fig:pareto}%
\end{figure}

The previous ablations show the BoNets performance with a ResNet-50 backbone and $224\times224$ image resolution. Here we study BoTNets when scaling up the model capacity and image resolution. There have been several works improving the performance of ConvNets on ImageNet~\cite{zhang2020resnest,tan2019efficientnet,bello2021revisit}. Bello \etal~\cite{bello2021revisit} recently propose scaling strategies that mainly increase model depths and increase the image resolutions much slower compared to the compound scaling rule proposed in EfficientNets~\cite{tan2019efficientnet}. We use similar scaling rules and design a family of BoTNets. The details of model depth and image resolutions are in the Appendix. We compare to the SENets baseline to understand the impact of the BoT blocks. The BoTNets and SENets experiments are performed under the same training settings (\eg, regularization and data augmentation). We additionally show EfficientNet and DeiT~\cite{touvron2021training} (regularized version of ViT~\cite{dosovitskiy2020image})\footnote{ViT refers to Vision Transformer~\cite{dosovitskiy2020image}, while DeiT refers to Data-Efficient Image Transformer~\cite{touvron2021training}. DeiT can be viewed as a regularized version of ViT with augmentations, better training hyperparameters tuned for ImageNet, and knowledge distillation~\cite{hinton2015distilling}. We do not compare to the distilled version of DeiT since it's an orthogonal axis of improvement applicable to all models.} 
to understand the performance of BoTNets compared with popular ConvNets and Transformer models. EfficientNets and DeiT are trained under strong data augmentation, model regularization, and long training schedules, similar to the training settings of BoTNets in the experiments.


{\bf ResNets and SENets are strong baselines until 83\% top-1 accuracy.}
ResNets and SENets achieve strong performance in the improved EfficientNet training setting. BoTNets T3 and T4 {\it do not} outperform SENets, while T5 does perform on par with S4. This suggests that pure convolutional models such as ResNets and SENets are still the best performing models until an accuracy regime of 83\%. 
{\bf BoTNets scale better beyond 83\% top-1 accuracy.}
While SENets are a powerful model class outperforms BoTNets (up to T4), we found gains to diminish beyond SE-350 (350 layer SENet described in Appendix) trained with image size 384. This model is referred to as S5 and achieves 83.8\% top-1 accuracy.
On the other hand, BoTNets scale well to larger image sizes (corroborating with our results in instance segmentation when the gains from self-attention were much more visible for larger images). In particular, T7 achieves 84.7\% top-1 acc., matching the accuracy of B7-RA, with a {\bf 1.64x} speedup in efficiency.
{BoTNets perform better than ViT-regularized (DeiT-384), showing the power of hybrid models that make use of both convolutions and self-attention compared to pure attention models on ImageNet-1K.} 

\section{Conclusion}
The design of vision backbone architectures that use self-attention is an exciting topic. We hope that our work helps in improving the understanding of architecture design in this space. Incorporating self-attention for other computer vision tasks such as keypoint detection~\cite{cao2017realtime} and 3D shape prediction~\cite{gkioxari2019mesh}; studying self-attention architectures for self-supervised learning in computer vision~\cite{henaff2019data, he2019momentum, tian2019contrastive, chen2020simple, grill2020bootstrap, chen2020exploring}; and scaling to much larger datasets such as JFT, YFCC and Instagram, are ripe avenues for future research. Comparing to, and incorporating alternatives to self-attention such as lambda-layers~\cite{bello2021lambdanetworks} is an important future direction as well.
\section{Acknowledgements}

We thank Ilija Radosavovic for several useful discussions; Pengchong Jin and Xianzhi Du for help with the TF Detection codebase; Irwan Bello, Barret Zoph,  Neil Houlsby, Alexey Dosovitskiy for feedback. We thank Zak Stone for extensive compute support throughout this project the through TFRC program providing Google Cloud TPUs (\url{https://www.tensorflow.org/tfrc}).



{\small
\bibliographystyle{ieee_fullname}
\bibliography{egbib}
}

\end{document}